\documentclass{article}

\usepackage{arxiv}
\usepackage[utf8]{inputenc} 
\usepackage[T1]{fontenc}    
\usepackage{hyperref}       
\usepackage{url}            
\usepackage{booktabs}       
\usepackage{amsfonts}       
\usepackage{nicefrac}       
\usepackage{microtype}      
\usepackage{lipsum}         
\usepackage{tcolorbox}
\usepackage{graphicx}
\usepackage{natbib}
\usepackage{doi}
\usepackage{mathbbol}
\usepackage{amssymb}
\usepackage{amsmath}
\usepackage{color}
\usepackage{tikz}
\usepackage{float}
\usetikzlibrary{arrows,shapes}
\usetikzlibrary{shapes.geometric}
\usetikzlibrary{arrows.meta}
\usepackage{graphics,graphicx}
\usepackage{mathrsfs}
\usepackage{amsmath,bm}
\usepackage{subfigure}
\usepackage{booktabs}
\usepackage{amsthm}
\usepackage{subfigure}
\usepackage{listings}
\usepackage{cleveref}       
\newtheorem{theorem}{Theorem}[section]
\newtheorem{corollary}{Corollary}[section]
\newtheorem{lemma}{Lemma}[section]

\newtheorem{definition}{Definition}[section]

\newtheorem{Remark}{Remark}[section]

\graphicspath{{Picture/}}
\title{
Numerical Approximation Capacity of Neural Networks with Bounded Parameters: Do Limits Exist, and How Can They Be Measured?
}

\newif\ifuniqueAffiliation{}
\uniqueAffiliationtrue{}

\ifuniqueAffiliation{} 

\author{ \href{https://orcid.org/0000-0000-0000-0000}{\includegraphics[scale=0.06]{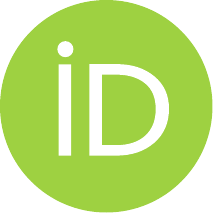}\hspace{1mm}Li Liu} \\
	Institute of Applied Physics and Computational Mathematics\\
	Beijing, CN, 100090\\
	\texttt{liu\_li@iapcm.ac.cn} \\
	\href{https://orcid.org/0000-0000-0000-0000}{\includegraphics[scale=0.06]{orcid.pdf}\hspace{1mm}Tengchao Yu}\\
	Institute of Applied Physics and Computational Mathematics\\
	Beijing, CN, 100090\\
	\href{https://orcid.org/0000-0000-0000-0000}{\includegraphics[scale=0.06]{orcid.pdf}\hspace{1mm}Heng Yong} \\
	Institute of Applied Physics and Computational Mathematics\\
	Beijing, CN, 100090\\
	\texttt{yong\_Heng@iapcm.ac.cn} \\
}
\else
\usepackage{authblk}

\newbox{\orcid}\sbox{\orcid}{\includegraphics[scale=0.06]{orcid.pdf}}
\author[1]{%
	\href{https://orcid.org/0000-0000-0000-0000}{\usebox{\orcid}\hspace{1mm}David S.~Hippocampus\thanks{\texttt{hippo@cs.cranberry-lemon.edu}}}%
}
\author[1,2]{%
	\href{https://orcid.org/0000-0000-0000-0000}{\usebox{\orcid}\hspace{1mm}Elias D.~Striatum\thanks{\texttt{stariate@ee.mount-sheikh.edu}}}%
}
\affil[1]{Department of Computer Science, Cranberry-Lemon University, Pittsburgh, PA 15213}
\affil[2]{Department of Electrical Engineering, Mount-Sheikh University, Santa Narimana, Levand}
\fi

\renewcommand{\shorttitle}{\textit{arXiv} Template}

\hypersetup{
pdftitle={A template for the arxiv style},
pdfsubject={q-bio.NC, q-bio.QM},
pdfauthor={David S.~Hippocampus, Elias D.~Striatum},
pdfkeywords={First keyword, Second keyword, More},
}

\begin{document}
\maketitle

\begin{abstract}

  The Universal Approximation Theorem posits that neural networks can theoretically possess unlimited approximation capacity with a suitable activation function and a freely chosen or trained set of parameters. However, a more practical scenario arises when these neural parameters, especially the nonlinear weights and biases, are bounded. This leads us to question: \textbf{Does the approximation capacity of a neural network remain universal, or does it have a limit when the parameters are practically bounded? And if it has a limit, how can it be measured?}

Our theoretical study indicates that while universal approximation is theoretically feasible, in practical numerical scenarios, Deep Neural Networks (DNNs) with any analytic activation functions (such as Tanh and Sigmoid) can only be approximated by a finite-dimensional vector space under a bounded nonlinear parameter space (NP space), whether in a continuous or discrete sense. Based on this study, we introduce the concepts of \textit{$\epsilon$ outer measure} and \textit{Numerical Span Dimension (NSdim)} to quantify the approximation capacity limit of a family of networks  both theoretically and practically.

Furthermore, drawing on our new theoretical study and adopting a fresh perspective, we strive to understand the relationship between back-propagation neural networks and random parameter networks (such as the Extreme Learning Machine (ELM)) with both finite and infinite width. We also aim to provide fresh insights into regularization, the trade-off between width and depth, parameter space, width redundancy, condensation, and other related important issues.
\end{abstract}


\section{Introduction}
Recalling the well-known universal approxmation theorem~\cite{hornik1989multilayer,pinkus1999approximation}:

\begin{theorem}\label{theorem1}
Let $g: \mathbb{R}^n\to \mathbb{R}$ be any non-polynomial function, and let $\mathbf{x} \in \mathcal{I}^n$, where $\mathcal{I}^n={[0,1]}^n$ is a n-dimensional unit cube. Then finite sums of the form
\begin{equation}\label{eq:1}
G(\mathbf{x}) = \sum_{j=1}^{\tilde{N}}\beta_j g(\mathbf{w}_j\cdot \mathbf{x}+b_j),\quad {\rm where} \quad \mathbf{w}_j \in \mathbb{R}^n, b_j, \beta_j \in \mathbb{R},
\end{equation}
are  dense in $C(\mathcal{I}^n)$.
In other words, given any $\varepsilon > 0$ and $f \in C(\mathcal{I}^n)$, $\exists \tilde{N} , \mathbf{w}_j, b_j$ and $\beta_j$ such that
\begin{equation}
||G(\mathbf{x}) - f(\mathbf{x})||<\varepsilon.
\end{equation}
\end{theorem}

Take the form Theorem~\ref{theorem1} as example, most of the research on the universal approximation property~\cite{chui1992approximation,hornik1989multilayer,pinkus1999approximation,cybenko1989approximation} is based on both the conditions that:
\begin{enumerate}
\item $\tilde{N} \in \mathbb{N}^+$ can be arbitrary large;
\item $\mathbf{w}_j\in \mathbb{R}^n$,  $b_j\in \mathbb{R}$ can be sampled or tuned without bound.
\end{enumerate}

However, both of these conditions are highly idealized in neural networks, so a more practical question arises: Does the Universal Approximation Theorem still hold when we restrict one or both conditions?

Some theoretical works promote and explore the approximation ability under bounded neuron width. \cite{guliyev2018approximation,guliyev2018approximation2} show that if the activation function \( g(\mathbf{x}) \) can be algorithmically constructed based on the interval of \( \mathbf{x} \), we can use only two neurons to approximate any continuous function when \( n=1 \). \cite{hanin2019universal} proves that for ReLU networks with arbitrary depth, there exists a minimum width that can approximate any continuous function on the unit cube \(\mathcal{I}^n = [0,1]^n\) arbitrarily well.

\subsection{Theoretical Approximation Ability with Bounded Parameter Space}
On the other hand, we consider the case where the  Nonlinear Parameters space (NPspace) of \((\mathbf{w}_j, b_j)\) are bounded. What will happen then?

The first study of universal approximation with bounded weights was provided by \cite{stinchcombe1990approximating}. They proved that in Eq.\ref{eq:1}, if \( g(\alpha) \) is a polygonal (piecewise linear) activation function with at least one and a finite number of kinks (such as ReLU), and if the weight bound \( B = \max(|\beta_j|, |\mathbf{w}_j|, |b_j|) \) is larger than \(\min_i \max(|\lambda_i - 2|, |\lambda_i + 2|)\), where \(\lambda_i\) is the \(x\)-coordinate of the kink point, then the polynomial spline function is proved to be universal under similar bound conditions as the polygonal function. This paper also analyzes that, for monotonic superanalytic functions, if the weight \( w_j \) is located on the unit sphere and \( b_j \) is bounded, \( G(x) \) can also achieve dense on any compact subsets of \( C(\mathbb{R}^n) \).

Adopting the idea of \( \mathbf{w}_j \) located on the unit sphere, \cite{ito1992approximation} carefully studied general sigmoid-type (may not be continuous or smooth) functions, giving many fruitful results about the universal approximation property. \cite{ismailov2012approximation,ismailov2015approximation,ismailov2017measure} vary \( \mathbf{w}_j \) on a finite set of straight lines and provide many theoretical conditions to achieve universal approximation under this condition. \cite{hahm2004approximation} gives a similar theorem that for bounded measurable sigmoid functions, there exist constants \( b_j, \delta_j \in \mathbb{R} \) and positive integers \( w_j = K \) and \( \tilde{N} \), such that any continuous function can be approximated arbitrarily well. \cite{maiorov1999lower} proves that there exists a real analytic, strictly increasing, and sigmoidal function \( g(\alpha) \) such that for a given bounded rigid function \( f(x) \), there exist constants \( \delta_j \), integers \( b_j \), and vectors \( \mathbf{w}_j \in \mathbf{S}^n \) such that \( f(x) \) can be arbitrarily approximated within \( \mathbf{x} \in \mathbf{B}^n \), where \( \mathbf{B}^n = \{\mathbf{x} : ||\mathbf{x}||_2 \le 1\} \) and \( \mathbf{S}^n = \{\mathbf{x} : ||\mathbf{x}||_2 = 1\} \).

Besides training the \( \mathbf{w} \) and \( b \) on the parameter space with a back-propagation (BP) of the target loss function, there is another kind of neural network method. The most representative is ELM and its derived methods, which use random generation and fixed \( \mathbf{w} \) and \( b \) on a given nonlinear parameter space. We call these methods random parameter (RP) neural networks. Compared with BP networks, RP networks converge more quickly because they only optimize the linear parameters \( \beta_j \).

In the construction of ELM, from both discrete and continuous perspectives, \cite{huang2006extreme} introduced a theoretical proof of the universal approximation property of ELM. Here, we refer to a discrete version of the approximation ability of the ELM method:

\begin{theorem}\label{theorem2}
Given any small positive value \( \varepsilon > 0 \), any activation function which is infinitely differentiable in any interval (meaning analytic, the restriction of "non-polynomial" is appended in the following quotes), and \( N \) arbitrary distinct samples \( (x_j, f_j) \in \mathbb{R}^n \times \mathbb{R}^m \), there exists \( \tilde{N} < N \) such that for any \( \{w_j, b_j\}_{i=1}^{\tilde{N}} \) randomly generated from any interval of \( \mathbb{R}^n \times \mathbb{R} \), according to any continuous probability distribution, with probability one, \( ||\mathbf{H}\beta - \mathbf{T}|| < \varepsilon \). And if \( \tilde{N} = N \), \( \mathbf{H} \) is invertible and \( \mathbf{H}\beta = \mathbf{T} \). Where \( \mathbf{H} \) is the \( \tilde{N} \times N \) matrix formed by \( [g(\mathbf{w}_j \cdot x_i + b_j)] \).
\end{theorem}

\subsection{The Gap Between Theory and Practice}
The referenced theorems theoretically guarantee that networks can approximate any function even when \( w \) and \( b \) are bounded in a finite nonlinear parameter space, both for BP neural networks \cite{stinchcombe1990approximating} and ELM \cite{huang2006extreme}. However, \cite{wang2011study} first discovered that the matrix \( \mathbf{H} \) is not full rank. Additionally, many studies show that ELM seems unable to approximate complex functions, even in low-dimensional space of \( \mathbf{x} \) with a small \( R \), which is the bound of \( w_j \) and \( b_j \). These problems lead to significant and difficult-to-understand deviations between theory and practice.

Here we consider a simple example:
\begin{tcolorbox}
\paragraph{Numerical Test:} We consider the following optimization problem:
\begin{equation}
  \min_{\beta_j} \left\| \sum_{i=1}^N \left( \sum_{j=1}^{\tilde{N}}\beta_j g(w_j x_i + b_j) - o_i \right) \right\|^2,
\end{equation}
where \( w_j, b_j \quad (j= 1,\cdots,\tilde{N}) \) are randomly sampled from the interval \([-1,1]\) based on a uniform distribution. \( x_i \quad (i=1,\cdots,N) \) are mesh-generated on the domain \([0,1]\) of the definition of the target function \( o(x) \), and \( o_i = o(x_i) \).

Here, we set
\[ o(x) = \sin(4\pi(x+0.05)) \cos(5\pi(x+0.05)) + 2, \]
and the activation function is set as \( g(x) = \text{Tanh}(x) \).
We choose \( N = \tilde{N} \) as 50, 100, 200, and 1000. The package \href{https://numpy.org/doc/stable/reference/routines.linalg.html}{numpy.linalg} is used to numerically solve the pseudo-inverse and the rank of the matrix \( \mathbf{H} \) (the TOL is set to the default \( 1 \times 10^{-12} \)). In Table~\ref{tab:1}, we present the \( L_2 \) and \( L_\infty \) errors of the results solved by ELM.

  \begin{table}[H]
    \centering
    \caption{Test of ELM}
    \label{tab:1}
    \begin{tabular}{|c|c|c|c|}
      \hline
      $N(\tilde{N})$ &  $L_2$     &  $L_\infty$  & $\text{rank}(\mathbf{H})$     \\ \hline
      50    &  0.046  &  0.1   &  14  \\ \hline
      100    & 0.018 &  0.038 & 14  \\  \hline
      200    & 0.017   & 0.035   & 15 \\ \hline
      1000    & 0.017 & 0.062 & 16 \\ \hline
    \end{tabular}
  \end{table}
\end{tcolorbox}

As shown in the above numerical test, although theoretically ELM can achieve universal approximation ability when \( N = \tilde{N} \), numerically, the rank of \( \mathbf{H} \) seems bounded, and we cannot increase accuracy by increasing the width \( \tilde{N} \) of the networks. This phenomenon also appears in BP networks, as noted by \cite{ismailov2017measure}, "But if weights are taken from too 'narrow' sets, then the universal approximation property is generally violated, and there arises the problem of identifying compact sets \( X \in  \mathbb{R}^d \) such that the considered network approximates arbitrarily well any given continuous function on \( X \)". Considering any small numerical tolerance, it seems that the  neural network has an approximation limit under a fixed and bounded nonlinear parameter space (NPspace).

\subsection{The topic of this paper}

Recently, with the development of scientific machine learning, neural networks are being used to solve partial differential equations (PDEs) \cite{}. In order to utilize the automatic differentiation (AD) process of neural networks to obtain the spatial/temporal derivatives of equations, smooth analytical activation functions have become more commonly used. Therefore, in this paper, we are interested in the following questions:

\textbf{Question 1.1:}
\label{Q1}
\emph{If both $\mathbf{w}_j$ and $b_j$ are bounded, considering any numerical tolerance $\epsilon$ and an analytic activation function $g(x)$, can the class $\{G(\mathbf{x})\}$ approximate any continuous functions, or does it have capability limits even with a large width $\tilde{N}$, and how can we measure it?}

This question is typically a problem in \textit{Measure Theory}. In Section 2, we begin our study by introducing a new \textit{Outer Measure} termed as 
\(\epsilon\)-Measure that considering tolerance in the measurement of subsets. We then concretize it to Euclidean space and function space.

As analyzing the family of functions in a given continuous space is very challenging, in Section 3, we focus on providing a theoretical study of the relationship between BP and RP networks. The conclusion shows that, under infinite width, BP networks and RP networks can approximate each other and both can achieve universal approximation. Therefore, we can use randomly sampled parameters to approximate the continuous NP space. Based on the study of RP networks, in Section 3, we prove from a continuous perspective that the 
\(\epsilon\)-Measure of a given network class is finite in any bounded NP space.

In Section 4, we provide a method for measuring 
\(\epsilon\)-Measure from a practical perspective. To quantify the finite capacity limit of the class
\(G(\mathbf{x})\), we introduce the concept of the \textit{Numerical Span dimension (NSdim)}. Additionally, numerical examples are used to show the NSdim of given neural networks in bounded parameter spaces by analyzing the \textit{Hidden Layer Output Matrix}. As shown in the numerical examples, we discuss the influence of width, depth, and the size of the parameter space on the NSdim.

To further utilize the theory of NSdim, we explore the relationship and differences between BP and RP neural networks and their respective advantages under finite width in Section 5.

At last of the introduction also detailed in the Conclusion, we will discuss the intention and emphasize the significance of this research.
First, from a macro perspective, with the development and widespread application of deep neural networks (DNNs), we now operate in a significantly different landscape compared to thirty years ago. Over the past three decades, theoretical research has primarily focused on exploring the universal approximation capabilities of neural networks under ideal conditions. However, the potential limitations in common settings have often been overlooked, leading to an overestimation of the network's approximation capabilities in engineering applications and algorithm design, frequently neglecting critical preconditions.

 \textbf{Numerical Tolerance:} While theoretical research on the universal approximation theorem remains fundamental, investigating the approximation properties of neural networks within a numerical context is crucial for their practical application as computational tools. Considering numerical tolerance (machine error) in the analysis of approximation capacity limits is essential, particularly when there is a significant gap between theory and practice. This is the primary contribution of this work.

\textbf{Parameter Space:} As Stinchcombe (1990) emphasized, "Bounded weights are necessary for any practical network implementation." Contrary to the original theory suggesting that any parameter space can facilitate universal approximation, this study highlights the importance of the Nonlinear Parameter Space (NPspace) for neural networks. It demonstrates how NPspace influences the network's approximation capacity limit, providing insight into why regularizations such as  $L_1$ and $L_2$
  are effective in simplifying networks. Additionally, this theory helps explain why the Extreme Learning Machine (ELM) method is particularly sensitive to parameter boundaries.

\textbf{Width Redundancy:} Numerically, with a given bounded NPspace (including post-training), increasing the width does not always enhance the complexity of the network. The width of the network can naturally lead to redundancy (linear correlation of neurons). Particularly when the number of neurons approaches or exceeds a certain threshold, more neurons will only result in correlation and redundancy. This threshold, termed as the \textit{Numerical Span Dimension (NSdim)}, can be used to measure the network's approximation capacity limit. This paper will demonstrate this concept and provide a method for approximating the NSdim.

\section{$\epsilon$ outer measure}
In the first section, we introduce a new \textit{outer measure} to assess a family of functions within the context of finite, non-infinitesimal numerical tolerance.
\begin{definition}
Let \( (M,d) \) be a metric space where \( M \) is a set and \( d \) is a metric on it, defined as:
\[ d: M \times M \to \mathbb{R}. \]

We consider a \textit{discrete subset} of \( M \), denoted \( S \subset M \), where each \( a \in S \) is an \textit{isolated point}. The collection of such subsets is denoted by \( \mathcal{S} \). We then propose a new definition of the \textit{sparsity of \( S_1 \)} from  \( S_2 \) as:
\[ \delta(S_1, S_2) = \sup_{a \in S_1}{} \inf_{b \in S_2}{} d(a, b). \]

Next,  with a given \(\epsilon \in \mathbb{R} \), we define the collection of \textit{\(\epsilon\)-sparsity subsets} of \( E \subset M \) as:
\[ \{ S \in \mathcal{S} : \delta(E, S) \leq \epsilon \}. \]

We can then define a new \textit{\(\epsilon\) outer measure}:
\begin{equation}
 \mu_\epsilon^*(E) = \inf_{S \in \mathcal{S}, \delta(E, S) \leq \epsilon} |S|.
\end{equation}
\end{definition}

In order to compare with the traditional definitions of Lebesgue measure and Radon measure, and to enhance understanding, we provide an alternative equivalent definition of the 
\(\epsilon\) outer measure.

\begin{definition}
For any point \( x \in M \) and a given real number \( \epsilon \), an \textit{\(\epsilon\)-Ball} is defined as:
\begin{equation}
   \bar{B}_\epsilon(x) = \{ y \in M : d(x, y) \leq \epsilon \}.
\end{equation}
We define \(\mathcal{B}\) as the collection of all \(\epsilon\)-Balls:
\begin{equation}
\mathcal{B} = \{ \bar{B}_\epsilon(x) : x \in M \}.
\end{equation}

Let \( E \subset M \). We define the \textit{\(\epsilon\) outer measure} as:
\begin{equation}\label{eq:def}
 \mu_\epsilon^*(E) = \inf_{\bigcup_{n=1}^\infty \bar{B}_n \supset E; \bar{B}_1, \cdots \in \mathcal{B}} \#\{\bar{B}_1, \bar{B}_2, \cdots\}. 
\end{equation}
\end{definition}
The \(\epsilon\) outer measure counts the smallest number of \(\epsilon\)-Balls that can cover all points of \( E \). It is somewhat similar to the construction of the Hausdorff measure, but here \(\epsilon\) is not infinitesimally small but 
 considered as finite with a given size.

Then we prove that \( \mu_\epsilon^*(\cdot) \) obeys the outer measure axioms:
\begin{enumerate}
\item  \textbf{Null empty set}: \( \mu_\epsilon^*(\emptyset) = 0 \). In fact, \( E = \emptyset \) is the necessary and sufficient condition for \( \mu_\epsilon^*(E) = 0 \).
\item  \textbf{Monotonicity}: If \( E_1 \subset E_2 \subset M \), then \( \mu_\epsilon^*(E_1) \leq \mu_\epsilon^*(E_2) \).
 
\begin{proof}
First, denote \(\mu_i= \mu_\epsilon^*(E_i)\) for simplicity, for all \(i \in \mathbb{N}^+ \).

Let $E_1\subset E_2$, and $E_2\subset \bigcup_{j=1}^{\mu_2} \bar{B}_j$, then  $ E_1 \subset \bigcup_{j=1}^{\mu_2} \bar{B}_j$, by the definition of (\ref{eq:def}), we have \(\mu_1 \le \mu_2\).

It is obvious that a collection of closed balls that can cover each \(E_i\) can surely cover their union. Therefore, the number of closed balls required to cover their union is no more than the sum of the number of closed balls needed to cover each \(E_i\).
\end{proof}

\item \textbf{Countable subadditivity}: If \(E_1, E_2, \cdots \subset M\) is a countable sequence of sets, then
\begin{equation}\label{eq:cs}
   \mu_\epsilon^*(\bigcup_{n=1}^\infty E_n) \leq \sum_{n=1}^\infty \mu_\epsilon^*(E_n).
\end{equation}
\begin{proof}
Considering \(\epsilon\) outer measure, it is meaningless to discuss subadditivity under infinite terms because only the empty set has zero measure. If \(\forall E_n \neq \emptyset\), then \(\sum_{n=1}^\infty \mu_\epsilon^*(E_n) = \infty\). Thus, (\ref{eq:cs}) always holds with infinite terms of \(E_n\). Therefore, we prove (\ref{eq:cs}) with finite terms. Without loss of generality, we prove
\[
\mu_\epsilon^*(E_1 \cup E_2) \le \mu_\epsilon^*(E_1) + \mu_\epsilon^*(E_2).
\]

Let \(\mu_1 = \mu_\epsilon^*(E_1)\) and \(\mu_2 = \mu_\epsilon^*(E_2)\). Then, 
\(E_1 \subset \bigcup_{j=1}^{\mu_1} \bar{B}_j^1\) and \(E_2 \subset \bigcup_{j=1}^{\mu_2} \bar{B}_j^2\).

Thus, 
\[E_1 \cup E_2 \subset \bigcup_{j=1}^{\mu_1 + \mu_2} \bar{B}_j,\]
where \(\bar{B}_j = \bar{B}_j^1\) for \(j = 1, \cdots, \mu_1\) and \(\bar{B}_{j+\mu_1} = \bar{B}_j^2\) for \(j = 1, \cdots, \mu_2\).

It is obvious that
\[\mu_\epsilon^*(E_1 \cup E_2) \le \mu_\epsilon^*( \bigcup_{j=1}^{\mu_1 + \mu_2} \bar{B}_j) \le \mu_1 + \mu_2.\]
\end{proof}

\end{enumerate}

\subsection{ In Euclidean Space}
Considering \( M = \mathbb{R}^n \), let us recall the definition of the Lebesgue outer measure:

\[
\mu_L^*(E) = \inf_{\bigcup_{i=1}^\infty H_i \supset E; H_1,\cdots \text{Boxes}} \sum_{i=1}^\infty \textbf{volume}(H_i)
\]

where \( H_i \) is an \( n \)-dimensional box of any size.

In Euclidean space, we can easily understand the meaning of the \(\epsilon\) outer measure: anything below the threshold of detection can be ignored.
Figuratively speaking, \( H_i \) can be considered as a ruler of arbitrary precision (known volume) for measuring other complex shapes. However, in practical calculations, we might only have a ruler with a certain level of roughness, can be seen as numerical tolerance. The \(\epsilon\) outer measure is designed for this purpose. Specifically, in physical space, when we want to measure the volume of an object, we usually ignore the impact of porosity on the volume. This is because our measurement precision cannot 'detect' the sizes of the voids, even though from the perspective of the Lebesgue measure, the total measure of the voids is not zero.

\subsection{ In Infinte dimensional vector space (Function Space)}\label{sec:fs}
In Euclidean space, the concept of $\epsilon$ outer measure is relatively straightforward, but we seek to address the problem of measure in function spaces. Beyond issues of tolerance, it is well-known that the Lebesgue measure cannot be extended to infinite-dimensional vector spaces. Let us consider the base set $M = C(X)$, which is the set of all continuous functions defined on $X$, where $X$ is a compact subset of $\mathbb{R}^n$. Equipped with a norm $\|\cdot\|$, inducing a metric, the closed ball $\bar{B}_\epsilon(\xi)$ is defined as:

\[
\bar{B}_\epsilon(\xi) = \left\{ f \in C(X) : \|f - \xi\| \leq \epsilon \right\},
\]

where $\xi \in C(X)$ and $\|\cdot\|$ is the norm on $C(X)$. This represents the set of all functions within $C(X)$ that lie within a distance $\epsilon$ from $\xi$ under the given norm.

Now, suppose $E \subset M$ is a family of functions, such as $E = \{ g(x;\theta) : \theta \in \mathbf{S}_\theta \}$, where $\mathbf{S}_\theta$ is a parameter space. We define $\mu_\epsilon(E)$ as the minimal number of functions needed to approximate every function in $E$ to within a tolerance of $\epsilon$. This can be interpreted as finding a (potentially finite-dimensional) subspace that sparsely approximates the infinite-dimensional subspace $E$ with a precision of $\epsilon$.

The introduction of $\mu^*_\epsilon$ allows us to compare two families of functions, even when both are infinite-dimensional. For instance, we can compare the following sets of functions:

\[
\left\{ g_1(\mathbf{w}_j \cdot \mathbf{x} + b_j) : (\mathbf{w}_j, b_j) \in \mathbf{S} \right\} \quad \text{vs.} \quad \left\{ g_2(\mathbf{w}_j \cdot \mathbf{x} + b_j) : (\mathbf{w}_j, b_j) \in \mathbf{S} \right\},
\]

or alternatively,

\[
\left\{ g(\mathbf{w}_j \cdot \mathbf{x} + b_j) : (\mathbf{w}_j, b_j) \in \mathbf{S}_1 \right\} \quad \text{vs.} \quad \left\{ g(\mathbf{w}_j \cdot \mathbf{x} + b_j) : (\mathbf{w}_j, b_j) \in \mathbf{S}_2 \right\},
\]

where $\mathbf{S}_1$ and $\mathbf{S}_2$ are different parameter spaces, or even between sets where both the functions $g(\cdot)$ and the parameter spaces $\mathbf{S}$ differ.

This transforms \textbf{Question 1.1} in the introduction into the following more precise mathematical problem:

For the family of functions

\[
\Xi(\mathbf{x} \in \mathbf{X} \subset \mathbb{R}^n; g \in C(\mathbf{X}), \mathbf{S} \subset \mathbb{R}^{n+1}) = \{g(\mathbf{x}; \theta) : \theta \in \mathbf{S} \},
\]

determine \(\mu^*_\epsilon(\Xi(\mathbf{x}, g, \mathbf{S}))\).

However, solving the \(\epsilon\) outer measure remains challenging, even with a given bounded NP-space \(\mathbf{S}\) within the scope of BP networks. Thanks to the introduction of ELM and other types of RP networks, we have a new approach to solving \(\mu^*_\epsilon(\Xi)\).

This involves analyzing the \textit{Hidden Layer Output Matrix}, which extends the concept of the \textit{Last Hidden Layer Output Matrix} from single-hidden layer feedforward neural networks (SLFNs) to multi-hidden layer feedforward neural networks (MLFNs).

But first, we need to theoretically address the relationship between BP and RP neural networks.

\section{BP neural networks can be approximated by RP neural networks}

%
We first given a definition of the Nonlinear Parameter Space:
\begin{definition}
Following Theorem~\ref{theorem1}, we aim to approximate any given continuous function $o(\mathbf{x}) \in C(\mathbb{R}^n)$ on any given compact set $\mathbf{X}\subset \mathbb{R}^n$ of $\mathbf{x}$.
Assume that weights $\mathbf{w}_j$ and biases $b_j$ are define in any subset $\mathbf{S}\subset \mathbb{R}^{n+1}$. Then we define the Nonlinear Parameter Space (NPspace) of the networks is $\mathbf{S}$, as the nonlinear parameters are $(\mathbf{w}_j,b_j)$. While $\mathbf{S}$ is compact, then the NPspace is bounded and closed.   Or we can loosely and simply refer to it as a bounded NPspace.
\end{definition}
By defining nonlinear parameters and the NPspace, we separate out the linear parameters. This is because optimizing linear parameters is much easier compared to nonlinear ones. Linear parameters are typically used solely as scaling transformations in the output layer, whereas nonlinear layers usually handle the fitting of the normalized feature space. Then we introduce the definition of universal approximator on given set.
\begin{definition}
Follow the definition in \cite{hornik1989multilayer}, consider a  measurable (usually Borel measurable) activation function $g(x):\mathbb{R}\to \mathbb{R}$  and
$G(\mathbf{x}) = \sum_{j=1}^{\tilde{N}}\beta_j g(\mathbf{w}_j\cdot \mathbf{x}+b_j)$ is a function mapping from $\mathbf{X}$ to  $\mathbb{R}$ are named as \textit{Single-hidden Layer Feedforward neural Networks (SLFN)},
if class $ \Pi^n(g)= \{G: G(\mathbf{x}) = \sum_{j=1}^{\tilde{N}}\beta_j g(\mathbf{w}_j\cdot \mathbf{x}+b_j): \tilde{N}\in \mathbb{N}^+,\theta_j=(\mathbf{w}_j,b_j) \in \mathbf{S}, \beta_j \in \mathbb{R}\}$ is dense in $\mathbf{C}(\mathbf{X})$, we say network $\Pi^n$ is a universal approximator on compact set $\mathbf{S}\times \mathbf{X} \subset \mathbb{R}^{2n+1}$.
\end{definition}
In another words, if $\Pi^n$ is {\bf a universal approximator} on $\mathbf{S}\times \mathbf{X}$
then for any given $f(\mathbf{x}) \in C(\mathbf{X})$ and $\varepsilon>0$,
there exists $\tilde{N}\in {\mathbb{N}}^+$, $\theta_j \in \mathbf{S}$,
and $\beta_j \in \mathbb{R}$  for every $j=1,2,\ldots,\tilde{N}$, such that
\begin{equation}
\left| f(\mathbf{x}) - G(\mathbf{x})\right|_\rho <\varepsilon,\quad \mathbf{x} \in \mathbf{X}.
\end{equation}


Before we begin discussing the approximation capability of $\Pi^n$, we introduce a theorem for the generalization of dimensions and intervals:
\begin{theorem}\label{theorem4}
Let $g(x): \mathbb{R}\to \mathbb{R}$ be a given measurable function. If $\Pi^1$ is dense on any given non-empty compact interval $x\in [-s,s]$ with bounded NPspace $\max(\theta_j) \in [-B,B]$ where $0 < B <+\infty$, then for every $n\in \mathbb{N}^+$, $\Pi^n$ on every compact subsets of $C(\mathbb{R}^n)$ with the same bound $B$ of $\max(\theta_j)$.
\end{theorem}
This theorem is directly deduced from Theorem 2.0 in~\cite{stinchcombe1990approximating} and it involves two levels of extension:
\begin{enumerate}
\item extending from input dimension $n=1$ to arbitrary higher dimensions,
\item extending from given closed intervals to arbitrary closed intervals.
\end{enumerate}
With the assistance of this theorem, our subsequent discussions only need to focus on the case of $n=1$ and a special given interval of $\theta$.
Theorems~\ref{theorem4} extends approximation limit from $n=1$ to every $n\ge 1$ and extends one given interval to each intervals.

 In this paper we mainly focus on the networks with analytic function with the definitions of:
\begin{definition}\label{def1}
  if measurable function $g(x): \mathbb{R}\to \mathbb{R}$  can be approximated by convegent serie $\sum_{i=1}^{\infty} c_i {(x-a)}^i$ for any $\{x:|x-a|< r\}$, then $g(x)$ is real analytic at $a\in \mathbb{R}$ with convergence radius of $r>0$. And if $c_n\ne 0$ for infinity terms of $n$, then $g(x)$ is superanalytic at $a$.
\end{definition}

After the definition of analytic function, we directly give a theoritical result of the approximation capability:
\begin{theorem}\label{theorem:theory}
If $g(x):\mathbb{R}\to \mathbb{R}$ is superanalytic and strictly monotonic then $\Pi^n(g)$ is universal approximator, if  $B \ge 1$.
\end{theorem}
The rigious proof is given in~\cite{stinchcombe1990approximating}. In what follows, let $\mathbf{S} = [-B,B]^{n+1}$ which is $n+1$-dimensional cube in $\mathbb{R}^{n+1}$.

\begin{theorem} \label{theorem:random}
Let $g(x) \in C(\mathbb{R})$, $\mathbf{S}$ is a cube in \( \mathbb{R}^{n+1}\), then families \( \Pi = \{G(\mathbf{x}): \tilde{N}\in \mathbb{N}^+, \theta \in \mathbf{S}, \beta_j \in \mathbb{R}\}\) and \(\Gamma = \{G(\mathbf{x}): \tilde{N}=+\infty, \theta \sim P, \beta_j \in \mathbb{R}\}\) can approximate each other with arbitrary precision with probability 1, for any given distribution \(P\) with positive probability to any subset of \(\mathbf{X}\) with positive Lebesgue measure.

\end{theorem}
\begin{proof}
By the difinition of continuous, if $g(x)$ is continuous on \(\mathbb{R}\), then $\forall x_0\in \mathbb{R}$, \(\forall \varepsilon >0\), \(\exists \delta = \delta(x_0,\varepsilon)>0\), that  \(\forall x: \left|x-x_0\right|\le \delta \), have \(|g(x)-g(x_0)|<\varepsilon\).

\(\Leftarrow\)
Consider a function  \(G_1(\mathbf{x})=\sum_{j=1}^{\tilde{N}}\beta_{0j} g(\mathbf{w}_{0j}\cdot \mathbf{x}_{j} + b_{0j}) \in \Pi\), \(\forall \varepsilon >0 \) and proof can find a function \(G_2(\mathbf{x}) \in \Gamma \) that
\[\sup_{\mathbf{x} \in \mathbf{x} \in \mathbf{X}}| G_1(\mathbf{x}) - G_2(\mathbf{x})|\le \varepsilon,\]
Due to the arbitrariness of \(\beta_j\) in  \(G_2\), we can always let
 \[\beta_j = \left\{
 \begin{aligned}
 &\beta_{j0}, &  \text{if}  \quad   1\le j\le \tilde{N},\\
 &0, & \text{if}  \quad  j > \tilde{N},\\
 \end{aligned} \right.
 \]
Then a sufficient condition for the validity of the above proposition is to prove that there must exist a term in \(G_2\) that can approximate \(g(\mathbf{w}_{0j}\cdot \mathbf{x}_j+b_{0j})\), with the required approximation accuracy of
\[\sup_{\mathbf{x}\in \mathbf{X}} | g(\mathbf{w}_{0j}\cdot \mathbf{x}+b_{0j})  - g(\mathbf{w}_{j}\cdot \mathbf{x}+b_{j}) | \le \varepsilon_j = \frac{1}{\tilde{N} \max(|\beta_j|)} \varepsilon. \]

Since the parameter spaces for \(\mathbf{x}\), \(\mathbf{w}_j\), and \(b_j\) are all compact, they are necessarily bounded, so \(|\mathbf{w}_j \cdot \mathbf{x}+b|\) must also be bounded. Let this bound be \(L\). Then, according to the definition of continuity, we can define \(\delta_{\text{min}} = \inf_{x \in [-L, L]} \delta(x, \varepsilon)\). To ensure the approximation holds, it is sufficient to satisfy \(|\mathbf{w}_j\cdot \mathbf{x}+b_j - \mathbf{w}_{0j}\cdot \mathbf{x} - b_{0j}| \leq \delta_{\text{min}}\). This can be achieved by constructing a \(n+1\)-dimensional cubic centered at \(\theta_{0j} = (\mathbf{w}_{0j}, b_{0j})\) with the lenth of \(d\) as: \(\text{cub}(\theta_{0j}, d)\), where the side length \(d\) only need to satisfy:
\[ d = \min_{i=1}^{n+1} d_i,\]
where 
\[d_i \le \left\{ \begin{aligned}
  &\frac{1}{n\sup_{\mathbf{X}}(|x_i|)} \delta_{\min} \quad & 1\le i\le n\\
  & \delta_{\min} \quad &  i=n+1
\end{aligned} \right.
  \]
As \(\text{cub}(\theta_{i0}, d)\) is a \(n+1\) dimensional cubic, then
\[\mu_L(\text{cub}(\theta_{j0},d)) = \mathbf{volume}(\text{cub}(\theta_{j0},d))
\]
Let \(p(\text{cub}(\theta_{j0},d)) = \eta >0\), as we only need one node located in \(\text{cub}(\theta_{j0},d)\) with propbility \(p \ge 1-\gamma\), \(\forall \gamma >0\), there must exisit an \(\tilde{N}< \infty\). When \(\tilde{N}\to \infty\), \(\gamma \to 0\).

\end{proof}
\begin{Remark}
Theorem~\ref{theorem:random} illustrates that, when the network is wide enough, as $\mathbf{w}_j$ and $b_j$ go dense on $\mathbf{S}_w$ and $\mathbf{S}_b$, the
nonlinear optimization problems can be transformed into linear optimization problems. This explains from the perspective of continuity why random and fix nonlinear parameter methods such as ELM and RFM are effective.
\end{Remark}

\begin{Remark}
Besides, there is another significant implication. We know that \(\Pi\)
 can easily achieve theoretical universality, and based on this theorem, it can be readily deduced that 
\(\Gamma\) is also universal.
\end{Remark}
\begin{Remark}\label{remark:1}
  There is a very important technique in this proof that we will repeatedly use later. In \(\Pi\), due to the influence of the linear coefficient \(\beta\), the problem becomes complex to analyze. However, the linear coefficient itself is not significant in the analysis, as linear combinations do not increase the dimensionality of the space. Moreover, practically, optimizing linear coefficients is simpler compared to nonlinear coefficients (typically involving linear least squares problems). Therefore, for various reasons, we can simplify the analysis of \(\Pi\) approximation ability to that of \(\Xi = \{g(\mathbf{w}\cdot \mathbf{x}+b): (\mathbf{w},b) \in \mathbf{S}, \mathbf{x} \in \mathbf{X}\}\). 
\end{Remark}

\begin{Remark}
Up to this point, all the discussions in this chapter have been theoretical, without considering the impact of any indivisible \(\epsilon\). However, when applying these theorems to real numerical problems, one must be very cautious. This is because the introduction of \(\epsilon\) may result in the network no longer being a universal approximator, and the absence of scale-invariance. Therefore, we introduce the following corollary to address this.
\end{Remark}

\begin{corollary}\label{co:1}
Let $\mathbf{S}$ is a compact subset in \( \mathbb{R}^{n+1}\) and \(\mathbf{X}\) is a compact subset in \(\mathbb{R}^n\), then for all \( \epsilon>0\) and  for all \(g(x) \in C(\mathbb{R})\), consider family  \(\Xi = \{g(\mathbf{w}\cdot \mathbf{x}+b): (\mathbf{w},b) \in \mathbf{S}, \mathbf{x} \in \mathbf{X}\}\), there exists a \(N \in \mathbb{N}\) \ that \(\mu_\epsilon^*(\Pi)\le N\).
\end{corollary}

\begin{proof}
Idea: For a given bounded parameter space, it can necessarily be covered by a finite number of \(\epsilon\)-balls. Considering that \(g(x)\) is a continuous function, for any  there exists \(\delta(\epsilon)\) from which an upper bound on the number \(N\) can be derived.

A formal proof will be presented in the published version of the article.
\end{proof}
For a given \(\epsilon\) that can be seen as numerical tolerance of a system, we can use \(\mu_\epsilon^*\) to estimate its approxiamtion ablility limit. But technically how?

\section{The Numerical Span Dimension of Neural Networks on a Discrete Space}
As pointed out in Remark \ref{remark:1}, the family of networks
\[
\Pi = \left\{ G(\mathbf{x}) = \sum_{j=1}^{\tilde{N}} \beta_j g(\mathbf{w}_j \cdot \mathbf{x} + b_j) : \mathbf{x} \in \mathbf{X}, \mathbf{w}_j, b_j \in \mathbf{S}, \beta_j \in \mathbb{R} \right\}
\]
is spanned by the function set
\[
\Xi(\mathbf{X}; g, \mathbf{S}) = \left\{ g(\mathbf{w}_j \cdot \mathbf{x} + b_j) : (\mathbf{w}_j, b_j) \in \mathbf{S} \right\}.
\]
If \(\Pi\) is to serve as a universal approximator, the dimension of \(\Xi\) must be infinite (this is why polynomial functions cannot be used as activation functions).

However, in practice, if we consider a radius \(\epsilon\) representing numerical tolerance, solving \(\mu_\epsilon^*(\Xi)\) involves finding a set of functions
\[
\{\varphi_1(\mathbf{x}), \varphi_2(\mathbf{x}), \dots, \varphi_\mu(\mathbf{x})\}
\]
such that for all \(\theta_j \in \mathbf{S}\), we have
\[
\inf_{i=1}^\mu \|g(\mathbf{x}; \theta_j) - \varphi_i(\mathbf{x})\| \leq \epsilon.
\]
Analogous to how measurable sets in Euclidean spaces can be approximated by a countable set of points, it follows from Theorem \ref{theorem:random} that the function family \(\Pi\) can also be approximated by discrete functions obtained through random sampling of parameters (or via a deterministic sequence using the Axiom of Choice (AC)). Consequently, the problem reduces to analyzing the dimension of the set
\[
\{ g(\mathbf{x}; \theta_1), g(\mathbf{x}; \theta_2), \dots, g(\mathbf{x}; \theta_\infty) \}.
\]
Similarly, we can approximate each function \(g(\mathbf{x}; \theta_j)\) with a countable set of discrete points \(\mathbf{x}_i \in \mathbf{X}\) under the assumption of the AC.

This transforms the problem into the study of the rank of the matrix
\[
\mathbf{H} =
\begin{bmatrix}
g(\mathbf{w}_1 \cdot \mathbf{x}_1 + b_1) & \cdots & g(\mathbf{w}_{\tilde{N}} \cdot \mathbf{x}_1 + b_{\tilde{N}}) \\
\vdots & \cdots & \vdots \\
g(\mathbf{w}_1 \cdot \mathbf{x}_N + b_1) & \cdots & g(\mathbf{w}_{\tilde{N}} \cdot \mathbf{x}_N + b_{\tilde{N}})
\end{bmatrix}_{N, \tilde{N} \to \infty}.
\]
Interestingly, this matrix \(\mathbf{H}\) corresponds exactly to the Last Hidden Layper Output Matrix used in the Extreme Learning Machine (ELM) method.

As:
\[
{\rm Dim}_{\Xi} = {\rm rank}_{N \to \infty, \tilde{N} \to \infty}(\mathbf{H}),
\]
this holds even when considering an approximation under \(\epsilon\), as defined by \(\mu_\epsilon^*(\Xi)\).

It is well known that the rank of \(\mathbf{H}\) is also equal to the number of non-zero singular values. According to matrix theory, analyzing the approximate rank of a matrix under a tolerance \(\epsilon\) requires examining the distribution of the matrix's singular values. The approximate rank is determined by counting the number of singular values that are significantly greater than a threshold \(\delta(\mathbf{H}, \epsilon)\).

Therefore, the properties of the singular values of \(\mathbf{H}\) directly reflect the properties of the network \(G(\mathbf{x})\). From a numerical perspective, we can define singular values greater than machine epsilon \(\epsilon\) as numerically non-zero singular values (NNSVs). We define the number of NNSVs as the \textbf{Numerical Span Dimension (NSDim)}. According to Theorem \ref{theorem:random}, both random parameter networks and backpropagation networks are ultimately constrained by the NSDim.

A necessary condition for any complex function to be numerically approximated by the span of \(\Xi\) is that the number of numerically non-zero singular values (NNSVs) grows without bound as the network width increases. Conversely, if the NSDim has an upper bound, then the expressive power of \(\Pi\) is also limited.

To explore this, we present the following lemma, theorem, and corollary for multi-layer neural networks:

\begin{lemma} 
Let
\[
\mathbf{H} = \left[
  \begin{array}{cccc}
  g(\mathbf{w}_1 \cdot \mathbf{x}_1 + b_1) & \cdots & g(\mathbf{w}_{\tilde{N}} \cdot \mathbf{x}_1 + b_{\tilde{N}}) \\
  \vdots & \ddots & \vdots \\
  g(\mathbf{w}_1 \cdot \mathbf{x}_N + b_1) & \cdots & g(\mathbf{w}_{\tilde{N}} \cdot \mathbf{x}_N + b_{\tilde{N}})
  \end{array}
  \right]_{N, \tilde{N} \to \infty},
\]
where \(\mathbf{w}_j\), \(b_j\), and \(\mathbf{x}_i\) are bounded. Then, the probability density distribution of the singular values of \(\mathbf{H}\) approximates a Dirac delta function \( \delta(0) \). That is, for any given \(\epsilon > 0\), the singular values of \(\mathbf{H}\) that are greater than \(\epsilon\) are bounded.
\end{lemma}

\begin{proof}
  Proof will be presented in the published version.
\end{proof}

\begin{theorem} 
\label{theorem:main}
If \(g(x) : \mathbb{R} \to \mathbb{R}\) is analytic and the parameters \((\mathbf{w}_j, b_j)\) are bounded, then for any \(N\) distinct nodes \(\mathbf{x}_i\) in a compact set \(\mathbf{S}_x \subset \mathbb{R}^n\), and considering any machine precision \(\epsilon > 0\), there exists \(M > 0\) such that
\[
{\rm NSDim}(\Pi) < M, \quad \text{for all } N, \tilde{N} \in \mathbb{N}^+.
\]
\end{theorem}

\begin{corollary}
For a neural network with \(L > 0\) hidden layers, let
\[
G^L(\mathbf{x}) = \sum_{j=1}^{\tilde{N}^L} \delta_j \circ_{l=1}^L \sigma_l(\mathbf{x}),
\]
where \(\sigma_l = g \circ A_l\), \(A_l(\mathbf{x}) = \mathbf{W}^l \cdot \mathbf{x} + \mathbf{b}^l\) is the affine transformation, \(\mathbf{W}^l\) is the weight matrix of size \(\tilde{N}^l \times \tilde{N}^{l-1}\), \(\tilde{N}^0 = m\) is the dimension of \(\mathbf{x}\), and \(\mathbf{b}^l\) is the bias vector of the \(l\)-th layer. The composition operator \(\circ_{l=1}^L \sigma_l\) denotes the composition of functions:
\[
\circ_{l=1}^L \sigma_l = \sigma_L \circ \cdots \circ \sigma_1.
\]
Then, the neural network dimension (\(\text{NSDim}\)) remains bounded for any number of layers.
\end{corollary}
\begin{Remark}
The above theorem asserts that, for any \(\epsilon > 0\), increasing the number of neurons or samples does not lead to the numerical approximation of all complex functions by \(\Pi\) and its span \(\Xi\). This implies that the neural network dimension (\(\text{NSDim}\)) is inherently bounded. However, a detailed analysis of the magnitude of \(\text{NSDim}\) remains open. Based on the lemma, numerical methods and matrix analysis tools can be used to study the properties of \(\mathbf{H}\) and gain insights into the NSDim.
\end{Remark}

\section{Numerical Tests}

Based on perception in Euclidean space, for instance, if a given \(\epsilon\) is considered, the number of \(\epsilon\)-balls that can cover a one-dimensional unit interval should roughly equal \(1/\epsilon\). When \(\epsilon\) is small, despite being finite, this still contains a substantial amount of information. However, for the function family \(\Xi\), the result is counter-intuitive.For analytic activation function, the NSDim is very small.

Below, we present few numerical tests with the common used activation functions ${\rm Tanh(\cdot)}$. Other analytic activation functions, such as \(\rm Sin(\cdot)\) and \(\rm Sigmoid(\cdot)\), exhibit similar conclusions.
The package scipy.linalg is used to handle the matrix computing. Without loss of generality, here we consider $n=1$ and the domain of $x$ is given as $[0, 1]$.
Here, $\varepsilon$ is set to $1\times 10^{-7}$ approaching the minimum values that can distinguish by single-precision floating-point number ($2^{-23}$). And $N$ is set equal to $\tilde{N}$ globally in the tests.
\subsection{Numerical Tests: NSDim .VS. Size of NPSpace and Width}

\begin{figure}
  \centering
  \includegraphics[width = 8cm]{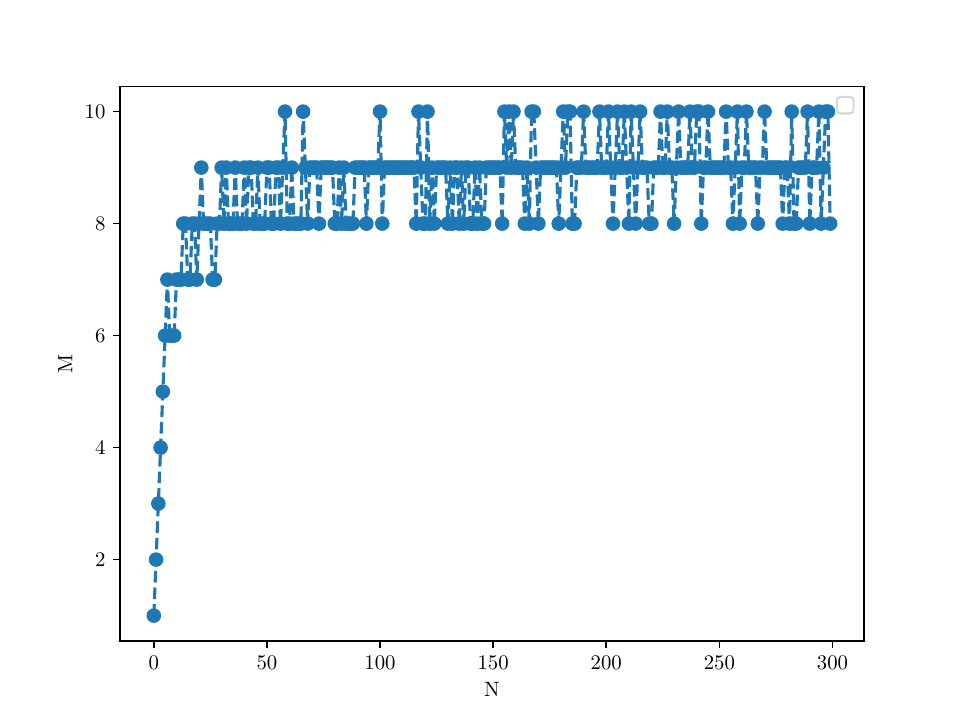}
  \includegraphics[width = 8cm]{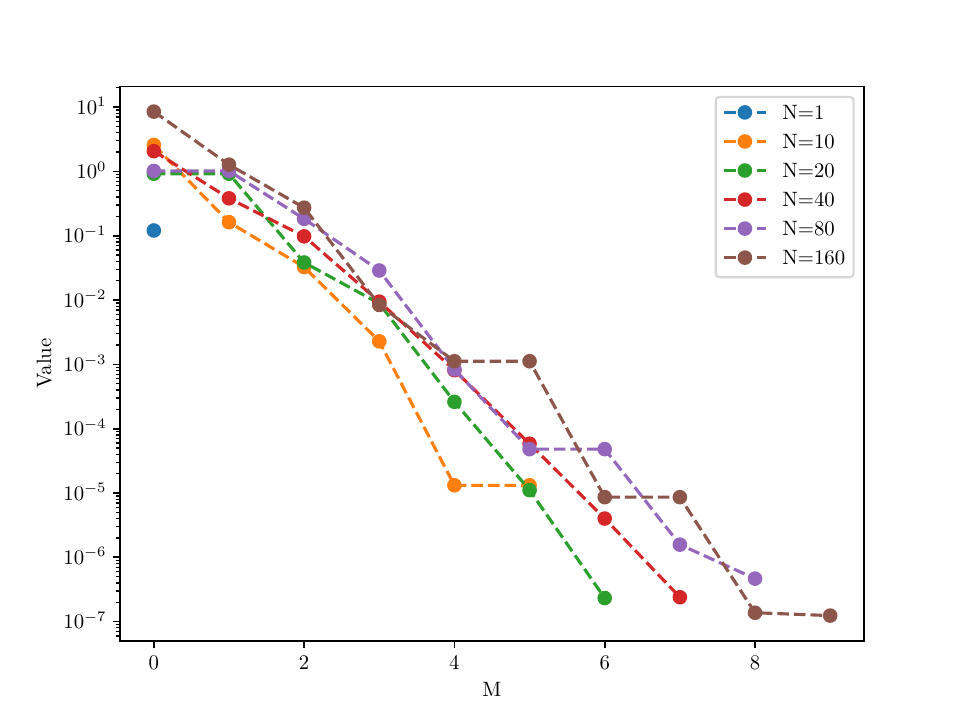}
  \includegraphics[width = 8cm]{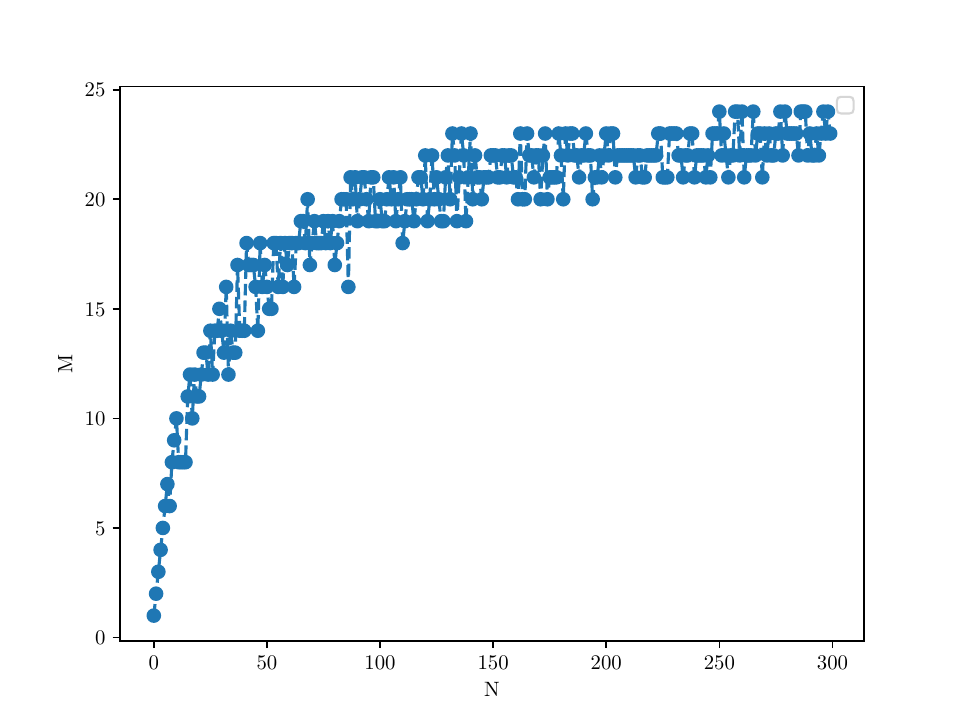}
  \includegraphics[width = 8cm]{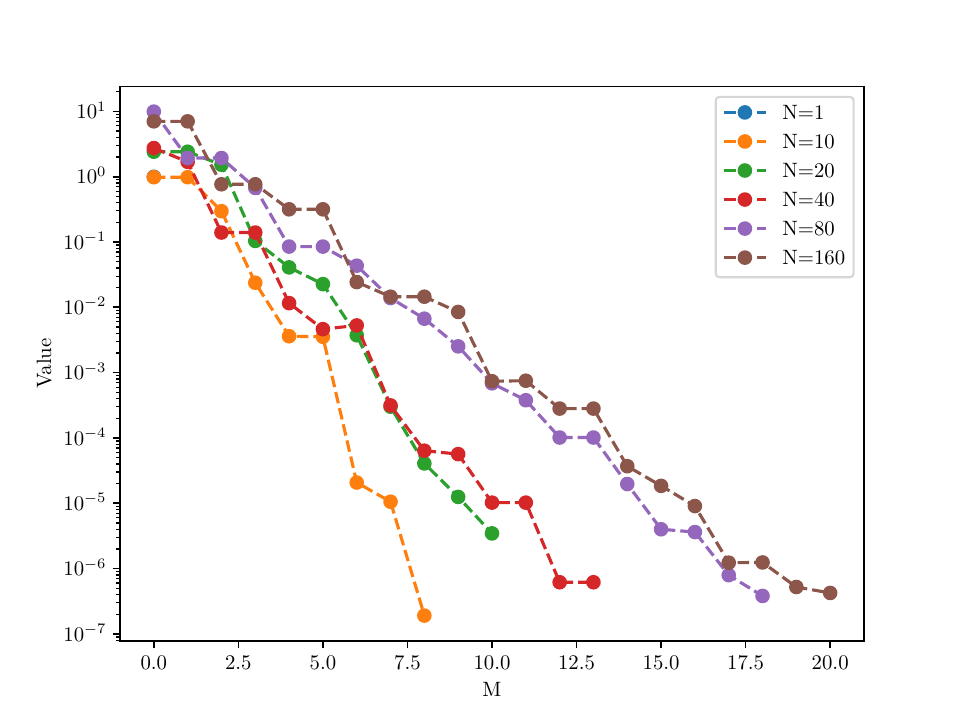}
  \includegraphics[width = 8cm]{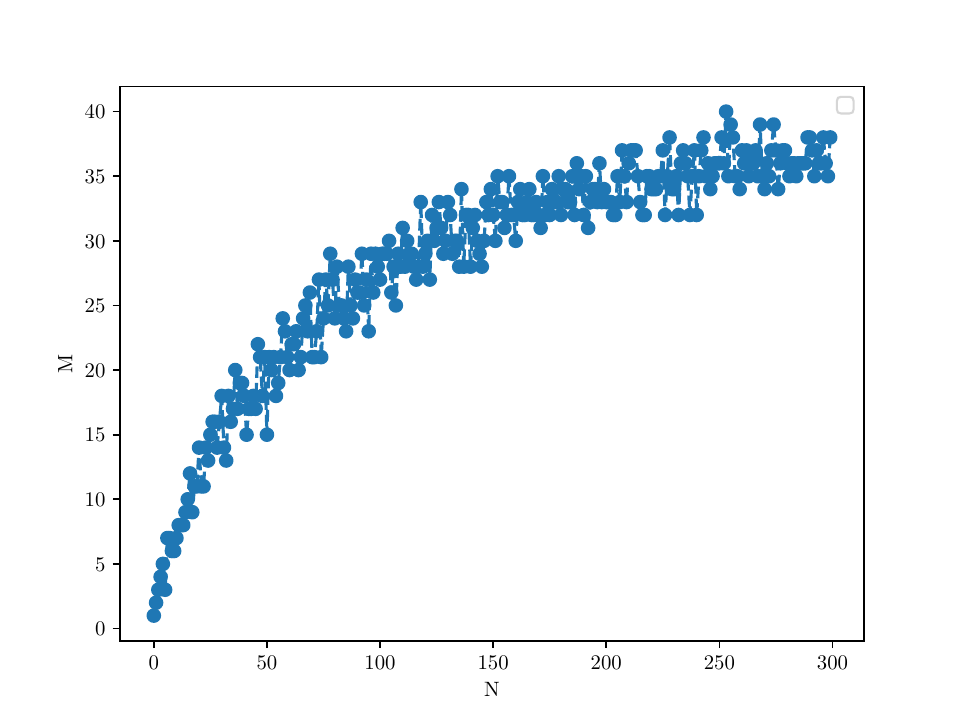}
  \includegraphics[width = 8cm]{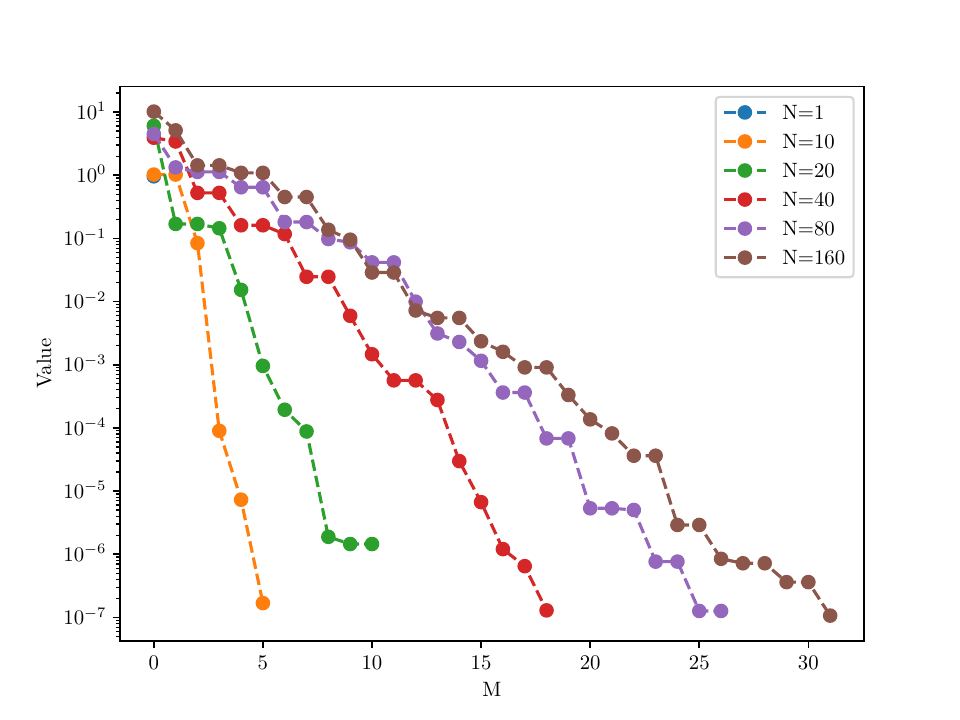}
	\caption{$\rm Tanh(\cdot)$ with random sampled of $x_i,w_j$ and $b_j$, $R=1,5$ and $10$ for different rows; Left: 
  the number of NNSVs with $N=\tilde{N}$; Right: NNSVs distribution. }~\label{fig:fig1}
\end{figure}

\begin{figure}
  \centering
  \includegraphics[width = 8cm]{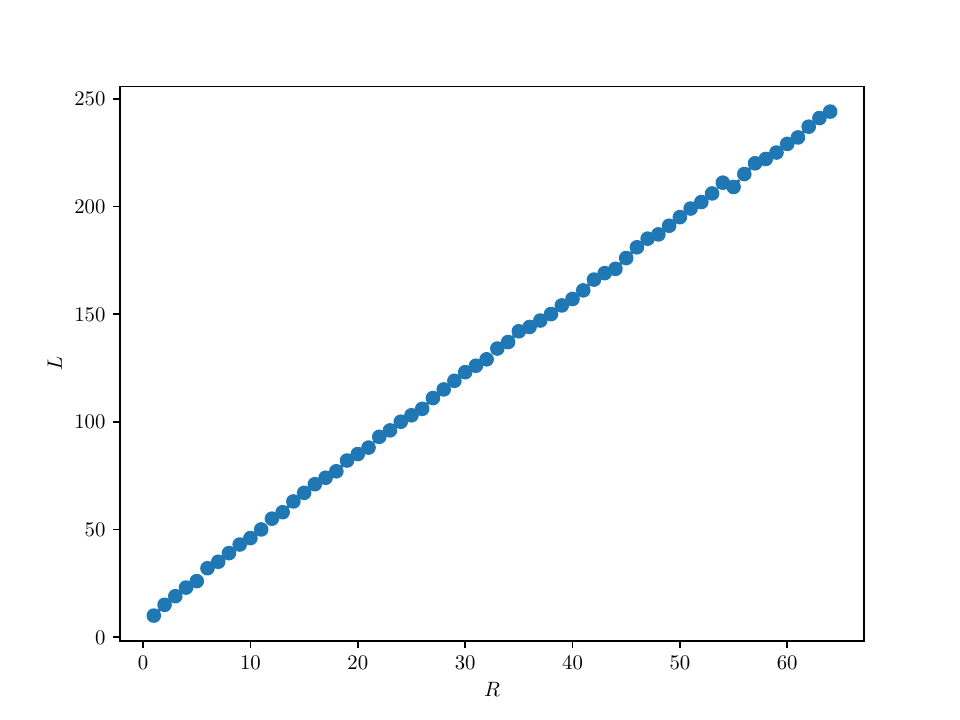}
	\caption{$\rm Tanh(\cdot)$, the relation of NSdim with $R$.}~\label{fig:fig4}
\end{figure}

First, we test $w_j$ and $b_j$ are randomly sampled based on uniform distribution on interval $[-R, R]$ ($R=1,5,10$) and $x_i$ is randomly sampled on the given interval $[0,1]$ with uniform distribution. In Fig.~\ref{fig:fig1}, we plot the relations bettween $N(=\tilde{N})$ and NSDim, also we shows the  coorespondingly NNSVs of matrix $\mathbf{H}$. In order to qualitatively understand the relations between the approximation limit NSDim and NPSpace size, in Fig.~\ref{fig:fig4}, we plot the relation of NSDim with $R$ (from 0 to 100). The results  illustrate that:
\begin{enumerate}
  \item  Mutually verified with theoretical conclusions, NSDim exsits and very small for all bounded $R$ which represents the size of NPSpace.
  \item NSDim is closed linear related to $R$.
  \item The singlar values of ${\mathbf H}$ decrease exponentially .
\end{enumerate}

\begin{Remark}
  We consider the impact of width of the network. As ${\rm NSDim}$ stands for the expression capacity limit,  when the width $\tilde{N}\ll {\rm NSDim}$, the expression capacity of the network grows linearly with $\tilde{N}$. However, when $\tilde{N}$ approaches or exceeds NSDim, a large number of neurons generate linear correlations.
  
  In fact,  neurons are inherently redundant, which stems from the bounded NSDim and the lack of orthogonality between the basis functions, even under good randomness. Even when the network width is still insufficient to match the NSDim, most neuron functions remain redundant. In BP networks, as training progresses based on the objective function, the randomness of the nonlinear parameters deteriorates, and the redundancy inevitably increases further. This may be a profound reason for the presence of {\bf Condensation Phenomena} in neural networks first finded by \cite{zhou2022towards, zhang2024implicit} that with small initialization, input weights of hidden neurons of neural networks will condense onto isolated orientations.
\end{Remark}

\subsection{Numerical Tests: NSDim .VS. Depth}

We have verified that the expressive capacity of neural networks is linearly correlated with the size of the parameter space. Specifically, given a fixed parameter space, when the network width is small, the expressive capacity grows linearly with the width. However, the increase in expressive capacity due to width is limited. As a result, we are interested in the impact of network depth.

In the following, we evaluate the expressive power of networks with one to three hidden layers. We measure the number of numerically non-zero singular values (NNSVs) for each layer, with a width of 10,000 in the given parameter space (\(R = 1\)). The number of NNSVs can be approximately regarded as the NSDim for different depths.

  \begin{table}[H]
    \centering
    \caption{Relationship between the numbers of NNSVs and hidden layers}
    \label{tab:2}
   \begin{tabular}{|c|c|}
      \hline
      Layers &  NNSVs  \\ \hline
      1    & 11   \\ \hline
      2    & 145   \\  \hline
      3    & 446   \\ \hline
      3    & 822   \\ \hline
    \end{tabular}
  \end{table}

  Tab.~\ref{tab:2} shows that as the deepth increasing, the NSDims increasing rapidly. It is thus evident that depth has a much stronger impact on the network's approximation capacity compared to width and NPspace.  It is important to note that due to the inherent redundancy of the network, the NSDim of networks with more than two hidden layers becomes difficult to investigate.
\section{BP  Neural Network .VS. RP Neural Network  under finite width}

As discussed above, for any bounded NPspace, there always exit ${\rm NSDim}<\infty$.

Then there must and only exist NSDim orthogonal basis functions
\begin{equation}
  \xi_j(\mathbf{x}), \text{ for } i=1,\cdots, {\rm NSDim}.
\end{equation}
And that within the tolerance of machine zero $\epsilon$, we have the approxiamtion of the two classes:
\begin{equation}
  \left\{ \xi_j(\mathbf{x}) \right\}  \to  \left\{ g(\mathbf{w}_j\cdot \mathbf{x} +b_j): \theta_j =(\mathbf{w}_j,  b_j) \sim P(\mathbf{S}) \right\}
\end{equation}

For a given network with finite width $\tilde{N}$, if the target function is given as $O(\mathbf{x})$, then the optimation problem of random parameter method can be given as
\begin{equation}(\mathcal{L}_1)\left\{
  \begin{aligned}
&\min_{\beta_j\in \mathbb{R}} \left \|\sum_{j=1}^{\tilde{N}} \beta_j g(\mathbf{w}_j\cdot \mathbf{x} +b_j) - O(\mathbf{x})\right\|_\infty \\
&  \mathbf{s.t.} \quad \theta_j \sim P(\mathbf{S}).
  \end{aligned}\right.
\end{equation}
Assume  
\begin{equation}
\dim\left\{ g(\mathbf{w}_j\cdot \mathbf{x} +b_j), (j=1,\cdots \tilde{N})  \right\} = {\rm NRP},
\end{equation}
then ofcause ${\rm NRP}\le \min(\tilde{N},{\rm NSdim})$.
And problem $\mathcal{L}_1$ has the same optimal solution with
\begin{equation}(\mathcal{L}_2)\left\{
  \begin{aligned}
&\min_{\beta_j\in \mathbb{R}} \left \|\sum_{j=1}^{\rm NRP} \beta_j \xi_j(\mathbf{x}) - O(\mathbf{x})\right\|_\infty \\
&  \mathbf{s.t.}.
  \end{aligned}\right.
\end{equation}

Then we consider the optimation problem of a back-propagation networks with finite width.
\begin{equation}(\mathcal{L}_3)\left\{
  \begin{aligned}
&\min_{\beta_j\in \mathbb{R}, \theta_j \in \mathbf{S}} \left \|\sum_{j=1}^{\tilde{N}} \beta_j g(\mathbf{w}_j\cdot \mathbf{x} +b_j) - O(\mathbf{x})\right\|_\infty \\
&  \mathbf{s.t.}
  \end{aligned}\right.
\end{equation}
it has same solution with
\begin{equation}(\mathcal{L}_4)\left\{
  \begin{aligned}
&\min_{\beta_j\in \mathbb{R}} \left \|\sum_{j=1}^{\rm NBP} \beta_j \eta_j(\mathbf{x}) - O(\mathbf{x})\right\|_\infty \\
&  \mathbf{s.t.}.
  \end{aligned}\right.
\end{equation}
where ${\rm NBP}=\min(\tilde{N},{\rm NSdim})$ and $\eta_j$ are the first $N_m$ principal basis functions of the projection from  $O(\mathbf{x})$  into the function space $\{\xi_j(\mathbf{x}),j=1,\cdots, {\rm NSdim}\}$.

Here we can compare problem $\mathcal{L}_1$($\mathcal{L}_2$) with $\mathcal{L}_3$($\mathcal{L}_4$) in different situation of the ralation between $\tilde{N}$ and NSDim.

\begin{enumerate}
  \item If $\tilde{N}  \gg  {\rm NSDim}$, then the optimal solution of $\mathcal{L}_2$ and $\mathcal{L}_4$ are all have same 
   solutions to the following linear optimization problem under a complete orthonormal basis:
  \begin{equation}
\left\{
  \begin{aligned}
&\min_{\beta_j\in \mathbb{R}} \left \|\sum_{j=1}^{\rm NSDim} \beta_j \xi_j(\mathbf{x}) - O(\mathbf{x})\right\|_\infty \\
&  \mathbf{s.t.}.
  \end{aligned}\right.
  \end{equation}
  However $\mathcal{L}_1$ is solving a linear problem while $\mathcal{L}_3$ is solving a nonlinear optimazation problem. Then $\mathcal{L}_1$ is better as it is easier to find the global optimal solution.
  \item If $\tilde{N}< {\rm NSDim}$, $\mathcal{L}_2$ can be understanded as random choose $\tilde{N}$ basis to optimazation, while $\mathcal{L}_3$ is to find $\tilde{N}$ `nearest' basis functions of $O(\mathbf{x})$ and optimizing linear parameters $\beta_j$. $\mathcal{L}_2$  has a probability of obtaining the same global optimal solution as the $\mathcal{L}_3$ problem, however, while $\tilde{N}\ll {\rm NSDim}$, it is more likely that optimizing only the linear coefficients can only approxiamte to the target with big error.
\end{enumerate}

\section{Conclusions and Talks}
The core contribution of this paper is the consideration of irreducible numerical tolerances on the approximation capacity of neural networks. We introduce a new outer measure and the concept of NSDim to quantify this effect, both theoretically and practically.
To achieve this goal, our works can be summarized as follows:
\begin{enumerate}
 \item  A new \(\epsilon\) outer measure is introduced to assess a family of functions within the context of ﬁnite, non-inﬁnitesimal numerical tolerance.
  \item  Theoretically, we proved the equivalence between random parameter (RP) networks, such as ELM, and back propagation (BP) networks when the network width tends to infinity, both possessing universal approximation capabilities. Therefore, we can analyze the network approximation capacity based on randomly parameterized networks.

  \item However, in a bounded {\bf Nonlinear Parameter Space~(NPspace)}, the infinite-dimensional space spanned by all the neural functions has only a limited number of dimensions that can be numerically utilized. This is prooved by the  analysis of the {\bf Hidden Layer Output Matrix} that only a finite number singular values are far aways from zero. Therefore, the infinite-dimensional space can be approximated by a finite-dimensional vector space. The dimensionality of this vector space, referred to as the {\bf Numerical Span Dimension (NSdim)}, can be used to measure the expressive capacity of the network.

  \item With the help of NSdim analysis, the theoretical reason why regularization works  (such as $L_1$ and $L_2$) can be easily explained. NSdim is positively correlated (nearly linearly) with the size of NPspace. By reducing NPspace, the complexity of the network is directly decreased, which enhances generalization capability. However, regrettably, it also affects the upper limit of network approximation ability.
  \item Numerically, it is hard to achieve full rank of Hidden Layer Output Matrices, and {\bf Numerical Non-zero Singular Values (NNSVs)} are even very sparse when the layer is wide enough,  and  the sparsity intensifies when the randomness of parameters deteriorates (e.g., due to  training). For this reason,  the width redundancy of neural networks is unavoidable, manifesting as a considerable amount of linear correlation among neurons. This may theoretically explain the occurrence of the phenomenon of coalescence, which is crucial for deepening our understanding of neural networks.

  \item  Till now, discussions about depth and width are primarily based on empirical studies. Based on our measurement of NSdim, we show that increasing depth is more advantageous than increasing width for increasing the upper limit of network complexity.

  \item Although the width naturally has redundancy, as we cannot require orthogonality among neuron functions, we can still determine whether the width has exceeded the upper limit by analyzing the NSdim of each layer. This is particularly necessary for the design of shallow networks or the first few layers of a deep network.
  \item We also provided analysis and commentary on the advantages and disadvantages of RP networks and BP networks based on NSdim analysis. The results indicate that when the target problem is relatively simple (corresponding to a low NSdim), RP methods have an absolute advantage due to the difference between linear and nonlinear optimization. However, when the target problem is complex (corresponding to a high NSdim), BP networks have the advantage.
\end{enumerate}

Certainly, due to the limitations of the author's expertise, this study has several known and unknown issues. The specific limitations and shortcomings identified include:
\begin{enumerate}
\item The findings in this paper primarily pertain to networks with analytic activation functions. For other commonly used non-analytic activation functions, certain conclusions still apply, such as inherent redundancy and behavior under finite width. However, conclusions regarding the smallness of NSDim may not hold universally and require further investigation.

\item This study largely abstracts away from the properties of the objective function, instead focusing on the upper bound of the expressive capacity of the function space itself.

\item Does a higher expressive capacity (larger NSDim) always imply better approximation capability? Clearly, this is not always right, the studies of regularization provides many examples. However, the drawbacks of increasing NSDim are an important open question. In future work, we will examine this issue more rigorously by analyzing the limitations in approximation accuracy within deep neural networks and exploring corresponding methods to address these challenges.
\end{enumerate}

\bibliographystyle{unsrtnat}
\bibliography{references}  

\begin{thebibliography}{18}
\providecommand{\natexlab}[1]{#1}
\providecommand{\url}[1]{\texttt{#1}}
\expandafter\ifx\csname urlstyle\endcsname\relax
  \providecommand{\doi}[1]{doi: #1}\else
  \providecommand{\doi}{doi: \begingroup \urlstyle{rm}\Url}\fi

\bibitem[Hornik et~al.(1989)Hornik, Stinchcombe, and
  White]{hornik1989multilayer}
Kurt Hornik, Maxwell Stinchcombe, and Halbert White.
\newblock Multilayer feedforward networks are universal approximators.
\newblock \emph{Neural networks}, 2\penalty0 (5):\penalty0 359--366, 1989.

\bibitem[Pinkus(1999)]{pinkus1999approximation}
Allan Pinkus.
\newblock Approximation theory of the {MLP} model in neural networks.
\newblock \emph{Acta numerica}, 8:\penalty0 143--195, 1999.

\bibitem[Chui and Li(1992)]{chui1992approximation}
Charles~K Chui and Xin Li.
\newblock Approximation by ridge functions and neural networks with one hidden
  layer.
\newblock \emph{Journal of Approximation Theory}, 70\penalty0 (2):\penalty0
  131--141, 1992.

\bibitem[Cybenko(1989)]{cybenko1989approximation}
George Cybenko.
\newblock Approximation by superpositions of a sigmoidal function.
\newblock \emph{Mathematics of control, signals and systems}, 2\penalty0
  (4):\penalty0 303--314, 1989.

\bibitem[Guliyev and Ismailov(2018{\natexlab{a}})]{guliyev2018approximation}
Namig~J Guliyev and Vugar~E Ismailov.
\newblock On the approximation by single hidden layer feedforward neural
  networks with fixed weights.
\newblock \emph{Neural Networks}, 98:\penalty0 296--304, 2018{\natexlab{a}}.

\bibitem[Guliyev and Ismailov(2018{\natexlab{b}})]{guliyev2018approximation2}
Namig~J Guliyev and Vugar~E Ismailov.
\newblock Approximation capability of two hidden layer feedforward neural
  networks with fixed weights.
\newblock \emph{Neurocomputing}, 316:\penalty0 262--269, 2018{\natexlab{b}}.

\bibitem[Hanin(2019)]{hanin2019universal}
Boris Hanin.
\newblock Universal function approximation by deep neural nets with bounded
  width and relu activations.
\newblock \emph{Mathematics}, 7\penalty0 (10):\penalty0 992, 2019.

\bibitem[Stinchcombe and White(1990)]{stinchcombe1990approximating}
Maxwell Stinchcombe and Halbert White.
\newblock Approximating and learning unknown mappings using multilayer
  feedforward networks with bounded weights.
\newblock In \emph{1990 IJCNN International Joint Conference on Neural
  Networks}, pages 7--16. IEEE, 1990.

\bibitem[Ito(1992)]{ito1992approximation}
Yoshifusa Ito.
\newblock Approximation of continuous functions on {R}d by linear combinations
  of shifted rotations of a sigmoid function with and without scaling.
\newblock \emph{Neural Networks}, 5\penalty0 (1):\penalty0 105--115, 1992.

\bibitem[Ismailov(2012)]{ismailov2012approximation}
Vugar~E Ismailov.
\newblock Approximation by neural networks with weights varying on a finite set
  of directions.
\newblock \emph{Journal of Mathematical Analysis and Applications},
  389\penalty0 (1):\penalty0 72--83, 2012.

\bibitem[Ismailov(2015)]{ismailov2015approximation}
Vugar~E Ismailov.
\newblock Approximation by ridge functions and neural networks with a bounded
  number of neurons.
\newblock \emph{Applicable Analysis}, 94\penalty0 (11):\penalty0 2245--2260,
  2015.

\bibitem[Ismailov and Savas(2017)]{ismailov2017measure}
Vugar~E Ismailov and Ekrem Savas.
\newblock Measure theoretic results for approximation by neural networks with
  limited weights.
\newblock \emph{Numerical Functional Analysis and Optimization}, 38\penalty0
  (7):\penalty0 819--830, 2017.

\bibitem[Hahm and Hong(2004)]{hahm2004approximation}
Nahmwoo Hahm and Bum~Il Hong.
\newblock An approximation by neural networkswith a fixed weight.
\newblock \emph{Computers \& Mathematics with Applications}, 47\penalty0
  (12):\penalty0 1897--1903, 2004.

\bibitem[Maiorov and Pinkus(1999)]{maiorov1999lower}
Vitaly Maiorov and Allan Pinkus.
\newblock Lower bounds for approximation by {MLP} neural networks.
\newblock \emph{Neurocomputing}, 25\penalty0 (1-3):\penalty0 81--91, 1999.

\bibitem[Huang et~al.(2006)Huang, Zhu, and Siew]{huang2006extreme}
Guang-Bin Huang, Qin-Yu Zhu, and Chee-Kheong Siew.
\newblock Extreme learning machine: theory and applications.
\newblock \emph{Neurocomputing}, 70\penalty0 (1-3):\penalty0 489--501, 2006.

\bibitem[Wang et~al.(2011)Wang, Cao, and Yuan]{wang2011study}
Yuguang Wang, Feilong Cao, and Yubo Yuan.
\newblock A study on effectiveness of extreme learning machine.
\newblock \emph{Neurocomputing}, 74\penalty0 (16):\penalty0 2483--2490, 2011.

\bibitem[Zhou et~al.(2022)Zhou, Qixuan, Luo, Zhang, and Xu]{zhou2022towards}
Hanxu Zhou, Zhou Qixuan, Tao Luo, Yaoyu Zhang, and Zhi-Qin Xu.
\newblock Towards understanding the condensation of neural networks at initial
  training.
\newblock \emph{Advances in Neural Information Processing Systems},
  35:\penalty0 2184--2196, 2022.

\bibitem[Zhang and Xu(2024)]{zhang2024implicit}
Zhongwang Zhang and Zhi-Qin~John Xu.
\newblock Implicit regularization of dropout.
\newblock \emph{{IEEE} Transactions on Pattern Analysis and Machine
  Intelligence}, 2024.

\end{thebibliography}
\end{document}